\def\year{2020}\relax
\documentclass[letterpaper]{article} % DO NOT CHANGE THIS
\usepackage{my_style}  % DO NOT CHANGE THIS
\usepackage{times}  % DO NOT CHANGE THIS
\usepackage{helvet} % DO NOT CHANGE THIS
\usepackage{courier}  % DO NOT CHANGE THIS
\usepackage[hyphens]{url}  % DO NOT CHANGE THIS
\usepackage{graphicx} % DO NOT CHANGE THIS
\usepackage{graphicx}  % DO NOT CHANGE THIS
\usepackage{siunitx}
\usepackage{array}
\usepackage{multirow}
\usepackage{makecell}
\usepackage{amsmath}
\usepackage{verbatim}
\usepackage{amsfonts}
\usepackage{graphicx}
\usepackage{color}
\usepackage{float}
\usepackage{bm}
\usepackage{arydshln}
\usepackage[noend]{algpseudocode}
\usepackage{tikz}
\usepackage{booktabs}
\usepackage[linesnumbered, ruled, vlined]{algorithm2e}
\newcommand{\citet}[1]{\citeauthor{#1} \shortcite{#1}} 
\newcommand{\citep}{\cite}

\title{Few-Shot Sequence Labeling with Label Dependency Transfer and Pair-wise Embedding}
\author{
	Yutai Hou\thanks{ Research Center for Social Computing and Information Retrieval, Harbin Institute of Technology, 
		email: {\tt \{ythou, yjliu, car, tliu\}@ir.hit.edu.cn}}
	\quad\quad
	\textbf{Zhihan Zhou}\thanks{ Department of Computer Science, Northwestern University, 
		email: {\tt \{zhihanzhou2020, ningwang2023\}@u.northwestern.edu, hanliu@northwestern.edu}}
	\quad\quad
	\textbf{Yijia Liu}\footnotemark[1]
	\quad\quad 
	\textbf{Ning Wang}\footnotemark[2] \\
	\quad\quad 
	\textbf{\Large Wanxiang Che}\footnotemark[1]
	\quad\quad 
	\textbf{\Large Han Liu}\footnotemark[2]
	\quad\quad 
	\textbf{\Large Ting Liu}\footnotemark[1] 
}
\begin{document}
	
	\maketitle
	
	\begin{abstract}
		% #################### version 4.0 #####################################
		While few-shot classification has been widely explored with similarity based methods, 
		few-shot sequence labeling poses a unique challenge as it also calls for modeling the label dependencies.
%		benefits from taking the dependencies between labels into account. 
		To consider both the item similarity and label dependency,
		we propose to leverage the conditional random fields (CRFs) in few-shot sequence labeling. 
		It calculates emission score with similarity based methods and obtains transition score with a specially designed transfer mechanism. 
		When applying CRF in the few-shot scenarios,  
		the discrepancy of label sets among different domains makes it hard to use 
		the label dependency learned in prior domains.
		To tackle this, 
		we introduce the \textit{dependency transfer} mechanism that transfers abstract label transition patterns. 
		In addition, 
		the similarity methods rely on the high quality sample representation,
		which is challenging for sequence labeling,
		because sense of a word
		is different when measuring its similarity to words in different sentences.
		To remedy this, we take advantage of recent contextual embedding technique, 
		and further propose a \textit{pair-wise embedder}.
		It provides additional certainty for word sense by 
		embedding query and support sentence pair-wisely.	
		Experimental results
		on slot tagging and named entity recognition show that
		our model significantly outperforms the strongest few-shot learning baseline by 
		11.76 (21.2\%$\uparrow$) and 12.18 (97.7\%$\uparrow$) F1 scores respectively in the one-shot setting. 

	\end{abstract} 
	
	\section{Introduction}
	% ################ version 2.0 ##############################
	Sequence labeling assigns a categorical label to each item of a sequence. 
	Many natural language processing (NLP) problems can be cast into sequence labeling (e.g., slot tagging and named entity recognition).
%	The state-of-the-art 
	Most
	sequence labeling models perform well 
	when the label set is fixed and sufficient training data is available. 
	But, sequence labeling problems, 
	such as slot tagging in task-oriented dialog system, 
	face the rapid change of domains and labeled data is usually scarce in a new domain. 
	Few-shot learning technique \citep{miller2000learning,fei2006one,lake2015human,matching} is appealing in this scenario since it learns a model that borrows the prior experience from the old (source) domains and adapts to the new (target) domains quickly even with very few labeled samples (usually one or two samples per class).
	
	\begin{figure}[t]
		\centering
		\begin{tikzpicture}
		\draw (0,0 ) node[inner sep=0] {\includegraphics[width=0.8\columnwidth, trim={11cm 8.9cm 13.2cm 4.3cm}, clip]{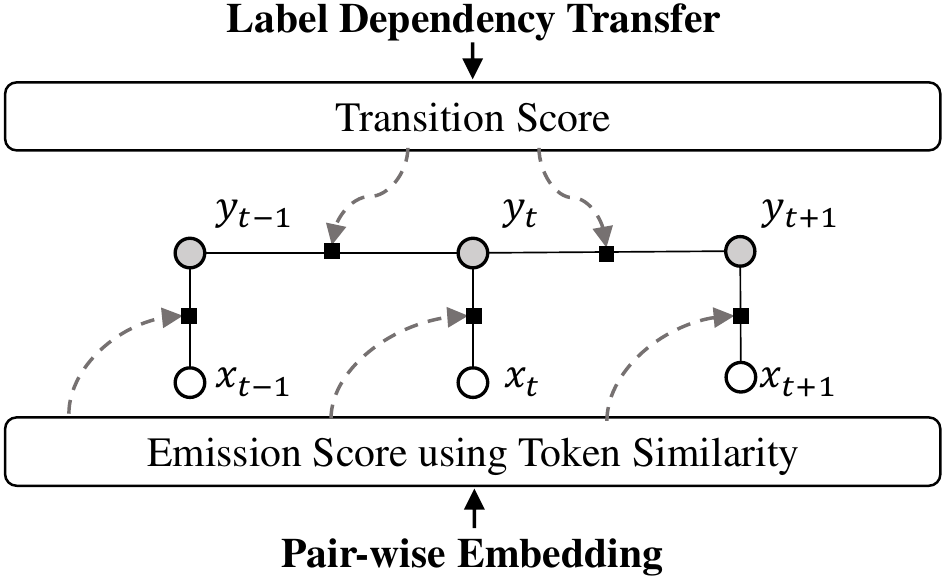}};
		\end{tikzpicture}  
		\caption{\footnotesize
			Linear-chain CRF for few-shot sequence labeling. 
			% 		 \wxcomment{no citation?}
		}\label{fig:intro1}
		\vspace*{-3mm}
	\end{figure}
	
	% For few-shot learning, 
	% the most studied prior experience is similarity metric. 
	Previous few-shot learning studies mainly focused on classification problems,
	which have been widely explored with similarity based methods \citep{matching,prototypical,sung2018learning,yan2018few,yu2018diverse}. 
	The basic idea of these methods is 
	classifying the items in new domain (query set) 
	according to their similarity with few labeled samples (support set)
	and the similarity function is usually learned in prior rich-resource domains.
	It is straight-forward to 
	decompose the few-shot sequence labeling into a series of independent few-shot classifications
	and apply similarity based methods.
	However, 
	sequence labeling benefits from taking the dependencies between labels into account \cite{huang2015bidirectional,ma2016end}.
	To consider both the item similarity and label dependency,
	we propose to leverage the conditional random fields (CRFs)
	 \citep{CRF} 
	 in few-shot sequence labeling (See Fig \ref{fig:intro1}).
	In this paper,
	we translate the emission score of CRFs into the output of the similarity based method
	and calculate the transition score with a specially designed transfer mechanism.
	
	The few-shot scenario poses unique challenges in learning the emission and transition scores of CRFs.
	It's infeasible to learn the transition on the few in-domain labeled data,
	and prior labeled dependency in source domain cannot be directly transfer
	due to discrepancy in label set.
	To tackle the label discrepancy problem, 
	we introduce the dependency transfer mechanism. 
	It transfers label dependency information from source domains to target domains 
	by abstracting domain specific labels into abstract labels and modeling the label dependencies between abstract labels.

	% It is straight-forward to 
	% decompose the few-shot sequence labeling into a series of independent \yjcomment{few-shot} classifications.
	% % and solve these few-shot classifications with the widely used similarity-based methods
	% % \citep{matching,prototypical,sung2018learning,yan2018few,yu2018diverse}. 
	% % The basic idea of these methods is classifying the input items (query set) according to their similarity with labeled samples (support set).
	% However, 
	% the success of these similarity-based methods 
	% relies on high quality representation that models the query-support interaction. 
	% This is challenging for sequence labeling due to ambiguity in word sense. 
	% As showed in Fig. \ref{fig:pair_emb}, 
	% a word tends to mean differently when concatenated to a different context.
	% More importantly,
	% % In addition, 
	% sequence labeling benefits from taking the dependencies between labels into account \citep{huang2015bidirectional,ma2016end}.
	% Thus, few-shot sequence labeling poses a unique challenge 
	% that it also calls for modeling the dependencies between labels.
	% % from the prior experience. 
	% Unfortunately, 
	% such label dependencies are hard to transfer from the source domains 
	% due to the label set discrepancy among different domains. 
	
\begin{figure}[t]
		\centering
		\begin{tikzpicture}
		\draw (0,0 ) node[inner sep=0] {\includegraphics[width=1\columnwidth, trim={2.5cm 4.85cm 3.1cm 3.5cm}, clip]{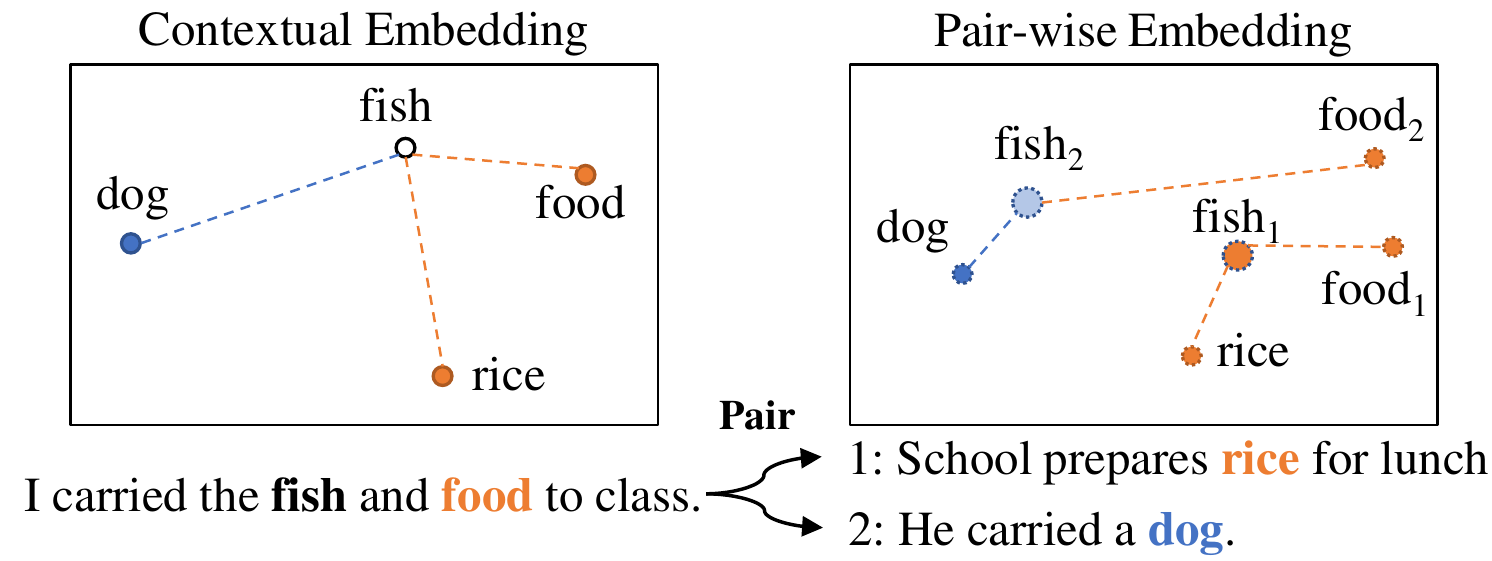}};
		\end{tikzpicture}
		\caption{\footnotesize
			An example of pair-wise embedding. 
			% 		 Due to co-occurrence, contextual embedding makes the \texttt{fish} closer to \texttt{food}. 
			% 		 But, after embedding pair-wisely, \texttt{fish}'s representation is close to \texttt{dog} as an animal.
			With single context (left), we can not decide whether \texttt{fish} is an animal or food.
			After pairing with different context (right), \texttt{fish} is more certain in meaning, and get close to \texttt{dog} and \texttt{rice} respectively. 
		}\label{fig:pair_emb}
%		\vspace*{-3mm}
	\end{figure}

	It is also challenging to calculate the emission scores
	(word-word similarity in our case) because
	the sense of a word differs when concatenated to a different context.
	As showed in Fig. \ref{fig:pair_emb}, 
	a word tends to mean differently.
	To tackle the representation challenges for similarity computation, 
	we consider the special query-support setting in few-shot learning 
	and suggest that embedding query and support tokens pair-wisely will provide additional certainty for word sense.
	Specifically, 
	we propose a pair-wise embedder that takes advantage of recent contextual embedding technique \cite{elmo,BERT},
	and represents a token with self-attention over both query and support sentence. 

	We perform few-shot experiments on two typical sequence-labeling tasks:
	slot tagging and named entity recognition, where label discrepancy problem is common.  
	Results show that our model
	achieves significant improvement over the few-shot learning baselines. 
	Ablation tests demonstrate improvements from both pair-wise embedding and label dependency. 
	%and yields more accurate sequence labeling results.
	%Ablation tests demonstrate that both pair-wise embedding and learnable scaler helps improve performance.
	Further analysis for label dependencies shows it captures non-trivial information and outperforms transition rules. 
	% and learnable scaler. 
	
	\begin{figure*}[t]
		\centering
		\begin{tikzpicture}
		\draw (0,0 ) node[inner sep=0] {\includegraphics[width=2.1\columnwidth, trim={5.1cm 6.3cm 11.05cm 3.5cm}, clip]{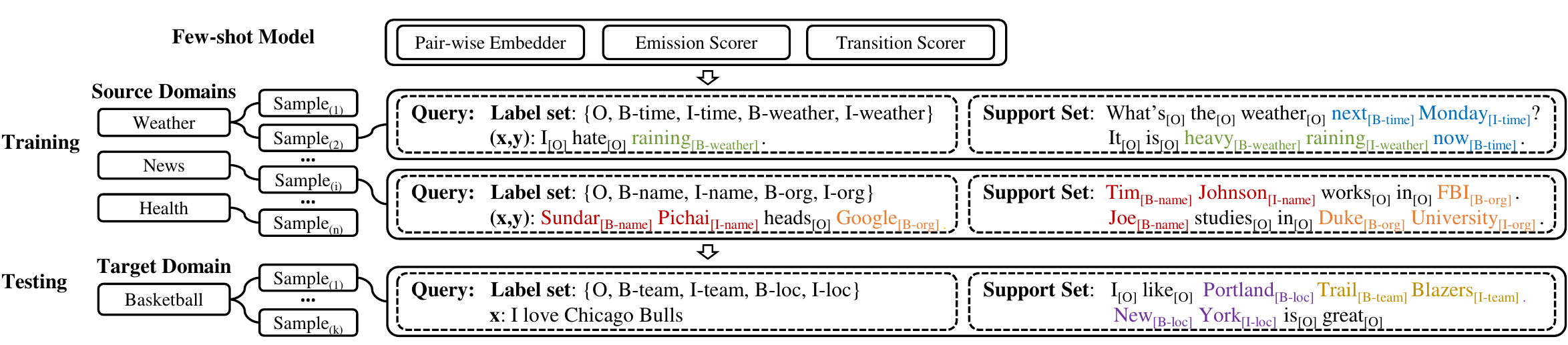}};
		\end{tikzpicture}
		\caption{\footnotesize
			Overviews of training and testing.
			This figure illustrates the procedure of training the model on a set of source domains, and testing it on an unseen domain with only a support set. }\label{fig:overview}
%		\vspace*{-3mm}
	\end{figure*}
	
	Our contributions are summarized as follows:
%	The main contributions of this work could be summarized as follows: 
	(1) We propose a CRF-based framework for few-shot sequence labeling, 
	which predicts labels by integrating the prior experience of token similarities and label dependencies.
	(2) We introduce the dependency transfer mechanism to transfer label dependencies across domains with different label sets. 
	(3) We leverage the special query-support structure in few-shot setting and introduce the pair-wise embedding to reduce representation ambiguity. 
	(4) Experiments show the superiority of our method over existing methods
	and emphasize the importance of transferring label dependencies and pair-wise embedding in few-shot sequence labeling.\footnote{Code and data will be available at: \url{https://github.com/AtmaHou/FewShotNLU}}

	\section{Problem Definition}\label{sec:p_def}
	%	The main difference between few-shot sequence labeling and conventional ones 
	%	is that few-shot models are designed to accurately label sequences from unseen domain with little labeled data.
	% Throughout this paper, we denote $\bm{X}=(X_1, X_2, \ldots, X_n)$ and $\bm{Y}=(Y_1, Y_2, \ldots, Y_n)$ as sequences of random variables, where $X_j$ corresponds to the $j_{th}$ token in the input sentence and $Y_j$ is the label of it. 
	% $x_j$ and $y_j$ are the realizations of them. 
	We define sentence $\bm{x} = (x_1, x_2, \ldots, x_n)$ as a sequence of tokens 
	and define label sequence of the sentence as $\bm{y} = (y_1, y_2, \ldots, y_n)$.
	A domain $\mathcal{D}  = \left\{(\bm{x}^{(i)},\bm{y}^{(i)})\right\}_{i=1}^{N_D}$ is a set of $(\bm{x},\bm{y})$ pairs. 
%	which is represented as $\mathcal{D}  = \left\{(\bm{x}^{(i)},\bm{y}^{(i)})\right\}_{i=1}^{N_D}$. 
	For each domain, 
	there is a corresponding domain-specific label set 
	$\mathcal{L_D} =\left\{ \ell_i \right\}_{i=1}^{N_\mathcal{L_D}}$. 
	% $\mathcal{L_D} = \left\{\ell_1, \ell_2, \ldots, \ell_{t_D} \right\}$.
	% And for each $(\bm{x}^{(i)},\bm{y}^{(i)})$ in $\mathcal{D}$, 
	% we assume $y^{(i)}_j \in \mathcal{L_D}$ holds for $j = (1,2, \ldots, n_i)$, 
	% where $n_i$ is the sequence length of $\bm{y}^{(i)}$. 
	
	\begin{algorithm}[t]
		\caption{Minimum-including}\label{algorithm}
		\footnotesize
		\begin{algorithmic}[1]
			\Require \# of shot $k$, domain $\mathcal{D}$, label set $\mathcal{L_D}$ \\
			Initialize support set $\mathcal{S}=\left\{ \right\}$, $\text{Count}_{\ell_j} = 0 $ 
			% 		$(i = 1, \ldots, N_\mathcal{LD})$
			$(\forall \ell_j \in \mathcal{L_D})$
			\\
			
			\For{$\ell$ in $\mathcal{L_D}$} {
				\While{$\text{Count}_{\ell} < k $ }
				{From $\mathcal{D} \setminus \mathcal{S}$, randomly sample a $(\bm{x}^{(i)},\bm{y}^{(i)})$ pair that $\bm{y}^{(i)}$ includes $\ell$
					
					Add $(\bm{x}^{(i)},\bm{y}^{(i)})$ to $\mathcal{S}$
					
					Update all $\text{Count}_{\ell_j}$ 
					% $(n = 1, 2, \ldots, t_D)$
					$(\forall \ell_j \in \mathcal{L_D})$
				}
			} \\
			
			\For{each $(\bm{x}^{(i)},\bm{y}^{(i)})$ in $\mathcal{S}$}
			{   
				Remove $(\bm{x}^{(i)},\bm{y}^{(i)})$ from $\mathcal{S}$ 
				
				Update all all $\text{Count}_{\ell_j}$ 
				% $(n = 1, 2, \ldots, t_D)$
				$(\forall \ell_j \in \mathcal{L_D})$ 
				
				\If{any $\text{Count}_{\ell_j} < $  k}
				{Put $(\bm{x}^{(i)},\bm{y}^{(i)})$ back to $\mathcal{S}$
					
					Update all $\text{Count}_{\ell_j}$ 
					% $(n = 1, 2, \ldots, t_D)$
					$(\forall \ell_j \in \mathcal{L_D})$
				}
			}\\
			
			Return $\mathcal{S}$ 
			
		\end{algorithmic}
		%\caption{Algorithm for randomly constructing the k-shot support set.}
	\end{algorithm}
	
	%Conventionally, researchers train and utilize model in the same domain. 
	%In this case, 
	%one could train a sequence labeling model for domain $D$ by splitting $\left\{(x^{(i)},y^{(i)})\right\}_{i=1}^{N_D}$ into train, valid and test sets. 
	%Then for each member $x_j$ of an input $x = (x_1, x_2, \cdots, x_n)$, the model assigns a label within $L_D$ to it. 
	As shown in Fig. \ref{fig:overview}, few-shot models are usually first trained on a set of source domains $\left\{\mathcal{D}_1, \mathcal{D}_2, \ldots \right\}$, 
	then directly work on another set of unseen target domains $\left\{\mathcal{D}_1', \mathcal{D}_2', \ldots \right\}$ 
	without fine-tuning. 
	% Here, $\mathcal{D}_i \neq \mathcal{D}_j'$ holds for any $i,j$. 
	% Target domain $\mathcal{D}_j'$ usually contains few $(\bm{x},\bm{y})$ pairs, which conforms to the few-shot setting. 
	A target domain $\mathcal{D}_j'$ only contains few labeled samples,
	which is called the support set $\mathcal{S} = \left\{(\bm{x}^{(i)},\bm{y}^{(i)})\right\}_{i=1}^{N_\mathcal{S}}$.
	$\mathcal{S}$ usually includes $k$ samples (\texttt{k-shot}) for each of $n$ classes (\texttt{n-way}).
	
	The k-shot sequence labeling task is defined as follows: 
	given a k-shot support set $\mathcal{S}$ and an input sequence $\bm{x} = (x_1, x_2, \ldots, x_n)$, 
	find $\bm{x}$'s best label sequence $\bm{y}^*$:
	\begin{equation} \nonumber
	\bm{y}^* = (y_1, y_2, \ldots, y_n) = \mathop{\arg\max}_{\bm{y}} \ \ p(\bm{y}  \mid  \bm{x}, \mathcal{S})
	\end{equation}
	
	%For an $n\text{-}way$ $k\text{-}shot$ classification task, 
	%we could built a support set by randomly picking n classes and sampling k different samples from each class.  
	For sequence labeling, 
	the normal \textit{n-way k-shot} definition is inapplicable.
	% an input $\bm{x}$ is a sequence of tokens, and each token has a label. 
	% These labels might be duplicate or not, 
	% so the number of labels within an input sequence is unpredictable. 
	% labels are assigned to tokens, 
	% but a support set consists of data samples of sentence level. 
	Because the sentence samples in a support set may include multiple duplicate token labels, 
	which makes the number of labels within a support set unpredictable.
	%To set a fair benchmark for researchers to compare their few-shot sequence labeling models, 
	We give the definition of \textit{k-shot} sequence labeling by introducing the \textit{k-shot} support set and presenting a minimum-including algorithm(Algorithm \ref{algorithm}) that randomly constructs a k-shot support set from original dataset. A k-shot support set $\mathcal{S}$ is essentially a set of $(\bm{x}^{(i)},\bm{y}^{(i)})$ pairs, each of which is sampled from the same domain $\mathcal{D}$. 
	A k-shot support set $\mathcal{S}$ follows two criteria: 
	(1) All labels within the domain should appear at least $k$ times in $\mathcal{S}$.  
	(2) At least one label will appear less than $k$ times in $\mathcal{S}$ if any $(\bm{x},\bm{y})$ pair is removed from it.

	%\vspace*{-3mm}
	\section{Model}\label{sec:model}
	In this section, we introduce the proposed framework. 
	The first subsection shows the overview of the framework, and we will respectively introduce the three components in the following three subsections. 
	%\ref{sec:trans}, \ref{sec:ctx}, \ref{sec:emission}. 
	
	\subsection{Framework Overview}\label{sec:overview}
	Conditional Random Field (CRF) considers both the transition score and the emission score to find the global optimal label sequence for each input.
	Following the same idea, we build our few-shot sequence labeling framework with three components: 
	Pair-wise Embedder, Emission Scorer and Transition Scorer. 
	%each of which can be implemented with different models.
	
	% We modify classic linear-CRF to conform it to the few-shot setting. 
	% Instead of modeling $p(\bm{y} \mid \bm{x})$, we model $p(\bm{x} \mid \bm{I})$. 
	% Here, $\bm{I} = (\bm{x}, \mathcal{S})$ and $\mathcal{S}$ is a k-shot support set. 
	% Then, we calculate the probability of $\bm{Y}=\bm{y}$ given $\bm{I}$ as:
	% \begin{equation} \nonumber
	% \begin{array}{rl}
	% p(\bm{y} \mid \bm{I})&= p(\bm{y} \mid \bm{x}, \mathcal{S}) \\
	% & = \frac{1}{Z}\exp(\text{TRANS}(\bm{y}) + \lambda \cdot \text{EMIT}(\bm{y}, \bm{x}, \bm{S})) \\
	% \end{array}
	% \end{equation}
	
	We apply the linear-CRF to the few-shot setting 
	by modeling the label probability of label $\bm{y}$ given query sentence $\bm{x}$ and a k-shot support set $\mathcal{S}$:
	\begin{equation} \nonumber
	p(\bm{y} \mid \bm{x}, \mathcal{S}) = \frac{1}{Z}\exp(\text{TRANS}(\bm{y}) + \lambda \cdot \text{EMIT}(\bm{y}, \bm{x}, \bm{S}))
	\end{equation}
	where
	\[
	Z = \sum_{\bm{y}' \in \bm{Y}}\exp (\text{TRANS}(\bm{y}') + \lambda \cdot \text{EMIT}(\bm{y}', \bm{x}, \bm{S}))
	\]
	$\text{EMIT}(\bm{y}, \bm{x}, \bm{S}) = \sum_{j=0}^n f_E(y_j, j, \bm{x}, \mathcal{S})$ is the Emission Scorer output 
	and $\text{TRANS}(\bm{y})=\sum_{j=1}^n f_T(y_{j-1}, y_j)$ is the Transition Scorer output. 
	$\lambda$ is a scaling parameter which balances weights of the two scores.
	Inspired by \citet{oreshkin2018tadam}, we learn it during training. 
	We take $-\log (p(\bm{y} \mid \bm{x}, \mathcal{S}))$ as loss function  
	and minimize it on data from source domains. 
	After the model is trained, we employ Viterbi algorithm 
	\citep{viterbi} to find the best label sequence for each input. 
	
	\begin{figure*}[t]
		\centering 
		\begin{tikzpicture}
		\draw (0,0 ) node[inner sep=0] {\includegraphics[width=1.95\columnwidth, trim={6cm 4.4cm 4.8cm 2.7cm}, clip]{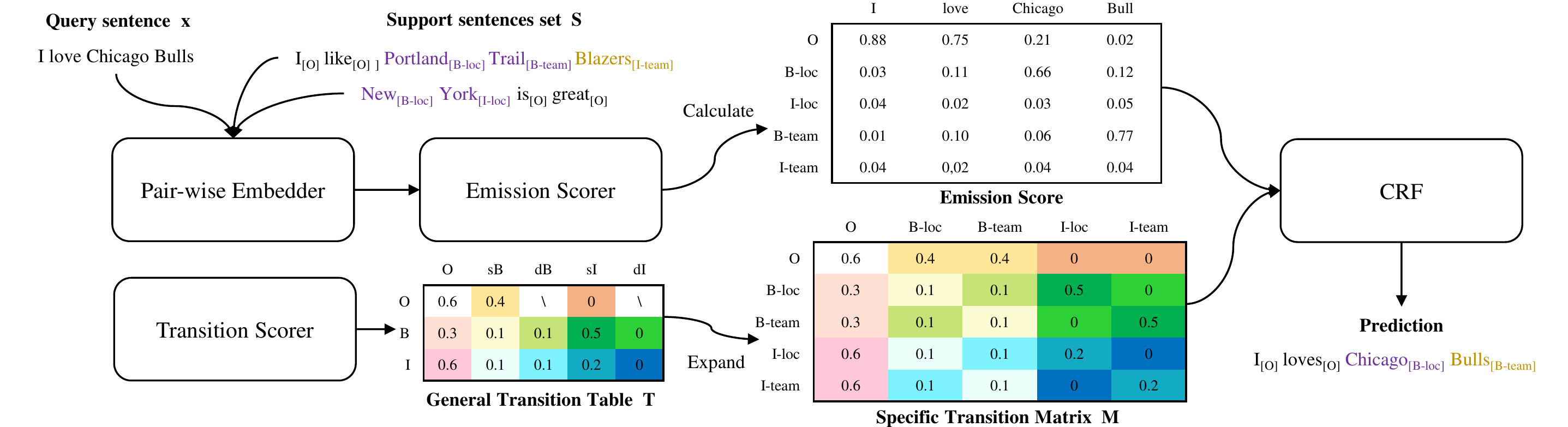}};
		\end{tikzpicture}
		\caption{\footnotesize
			Predicting procedure overview.
			This figure shows how each component performs during a single prediction. 
			In the right part, each color stands for a certain type of transition probability. 
			For each position of $\bm{M}$, we fill in it with value in $\bm{T}$ that has the same color.
			%We construct Transition Matrix \textit{M} for a new domain by filling in each position
			%with the values in Table \textit{T}. 
			%The rule of built an transition matrix \textit{M} with Table \textit{T} is shown in the right part, where 
		}\label{fig:framework}
	\vspace*{-1mm}
	\end{figure*}
	
	\subsection{Transition Scorer}\label{sec:trans}
	The transition scorer component captures the dependencies between labels.\footnote{Here, we ignore $Start$ and $End$ labels for simplicity.} 
	We model the label dependency as the transition probability between two labels:
	% \begin{equation} \nonumber
	% f_T(y_{j-1}, y_j) = p(Y_{j} = y_j \mid Y_{j-1} = y_{j-1} ) 
	% \end{equation}
	\begin{equation} \nonumber
	f_T(y_{j-1}, y_j) = p(y_j \mid y_{j-1} ) 
	\end{equation}
	
	Conventionally, such probabilities are learned from training data and stored in a transition matrix $\bm{M}^{N \times N}$, where $N$ is the number of labels. 
	% We define the first dimension to represent the rows and the second dimension to represents the columns.
	For example, $\bm{M}_{B\text{-}{\ell_1},B\text{-}{\ell_2}}$ corresponds to 
	% $p(Y_{j} = B\text{-}{\ell_2} \mid Y_{j-1} = B\text{-}{\ell_1})$, 
	$p(B\text{-}{\ell_2} \mid B\text{-}{\ell_1})$, 
	where $\ell_1$ and $\ell_2$ are two different labels.
	But in the few-shot setting, a model faces different label sets 
	in the source domains (train) and the target domains (test). 
	This mismatch on labels blocks the trained transition scorer
	directly working on a target domain. 
	%We therefore present the dependency transfer mechanism (DT) to overcome such issue.
	%The implementation of dependency transfer mechanism includes two stages.
	
	%The first stage occurs in training phase (on source domains).
	The dependency transfer mechanism (DT) overcomes this issue by %encoding the transition probability between real labels into 
	directly modeling the transition probabilities between abstract labels.
	Intuitively, there are only three abstract labels:  $O$, $B$ and $I$. 
	However, having merely $B$ and $I$ is not enough since the transition probability between the same labels and different labels are usually different.
	For example, $I$-$\ell_1$ is very likely to transit to $I$-$\ell_1$ itself, but will never transit to $I$ of another different label such as $I$-$\ell_2$.
	Therefore, we introduce another 4 abstract labels named same $B$ ($sB$), different $B$ ($dB$), same $I$ ($sI$) and different $I$ ($dI$).
	Then, instead of learning a transition matrix $\bm{M}$ for the real labels, 
	we learn a transition Table $\bm{T}^{3\times5}$ (see Fig. \ref{fig:framework}). %together with two $3$ dimensional lists: $L^S$ and $L^E$. 
	%Each row of $T$ respectively corresponds to $O$, $B$ and $I$. 
	%In addition, each of the column respectively corresponds to $O$, $sB$, $dB$, $sI$, and $dI$, respectively. 
	
	%Each position of $L^S$ respectively records the probability of transition from $Start$ token to $O$, $B$ and $I$. List $L^E$, on the other hand, records the probability of transition from $O$, $B$ and $I$ to $End$ token.
	% Each row of $\bm{T}$ respectively records the probabilities of transition from abstract labels $O$, $B$ and $I$ to abstract labels
	% $O$, $sB$, $dB$, $sI$, and $dI$. 
	$\bm{T}$ records the probabilities of transition from abstract labels $O$, $B$ and $I$ to abstract labels
	$O$, $sB$, $dB$, $sI$, and $dI$. 
	For example, $\bm{T}_{B,sB}$ stands for the probability of transition from $B$ of one label to $B$ of the same label, 
	which equals to 
	% $p(Y_{j} = B \text{-} {\ell_m} \mid Y_{j-1} = B\text{-}{\ell_m} )$. 
	$p(B \text{-} {\ell_m} \mid B\text{-}{\ell_m} )$. 
	$\bm{T}_{B,dI}$ stands for the probability of transition from $B$ of one label to $I$ of another different label,
	% We represents such probability as 
	which is 
	$p(I\text{-}{\ell_n} \mid B\text{-}{\ell_m} )$, 
	where $\ell_m \neq \ell_n$. 
	Also, $\bm{T}_{O,sB}$ and $\bm{T}_{O,sI}$ respectively stands for 
	the probability of transition from $O$ to any $B$ and $I$ label. 
	$\bm{T}_{O,dB}$ and $\bm{T}_{O,dI}$ are not used. 
	% In summary, transition table $\bm{T}$ records the transition probabilities between abstract labels.
	
	To calculate the label transition probability for a new domain, 
	we construct the transition matrix $\bm{M}$ by filling it with values in $\bm{T}$.
	For example, if there are only two labels named $\ell_1$ and $\ell_2$ in the new domain, we fill $\bm{M}_{B\text{-}{\ell_1},B\text{-}{\ell_2}}$ and $\bm{M}_{B\text{-}{\ell_2},B\text{-}{\ell_1}}$
	with value in $\bm{T}_{B,dB}$, and fill $\bm{M}_{B\text{-}{\ell_1},B\text{-}{\ell_1}}$ as well as $\bm{M}_{B\text{-}{\ell_2},B\text{-}{\ell_2}}$ with value in $\bm{T}_{B,sB}$.
	Fig. \ref{fig:framework} shows an example of table filling,
	where positions in the same color are filled by the same values.
	
	\subsection{Pair-wise Embedder}\label{sec:ctx}
	%Since we assign each token a label based on its representation (embedding)
	%and a token might have different meanings among different contexts, 
	%there is a natural demand that we should represent a token by taking its context into account.
	%Here, it is achieved with contextual embedding methods \citep{elmo,BERT} that embed  
	%each token based on both the token itself and its context. 
	%%	By taking the context into account, such embedding can effectively capture the dependencies between tokens. 
	%The Contextual Embedding component here takes a sequence of tokens $\bm{x} = (x_1, x_2, \ldots, x_n)$
	%as input, and output an embedding matrix $E^{n \times h}$, where $h$ 
	%is the embedding size. The $i_{th}$ row in this embedding matrix is the contextual word embedding of the $i_{th}$ token in $x$. 
	%%	In our implementation, 
	%%	contextual word embedding is given by a pre-train BERT model \citep{BERT}.
	Pair-wise embedder prepares the token representation for query utterance $\bm{x}$ and support set $\mathcal{S}$. 
	
	Since we assign each token a label based on its representation (embedding)
	and a token might have different meanings within different contexts, 
	there is a natural demand that we should represent a token by taking its context into account. 
	Furthermore, a token in different query-support pair should have different representation. 
	
	To achieve this, 
	we leverage pair-wise self-attention that represents each token with self-attention over both query and support tokens. 
	Pair-wise attention first copies query sentence $\bm{x}$ for $N_\mathcal{S}=|\mathcal{S}|$ times, 
	and pairs it with all support sentences.
	Then the $N_\mathcal{S}$ pairs are passed to a Transformer \citep{NIPS2017_7181} to get $N_\mathcal{S}$ paired representation sequences. 
	Now, tokens in query sentence have context-aware embedding and are represented differently for different support sentence.  
	Notice that tokens in support set will also have different representation for different query sample. 
	
	In our implementation, 
	we fine-tune pre-trained BERT \citep{BERT} as Transformer model, 
	since it is naturally capable of capturing relation between sentence pair. 
	
	\subsection{Emission Scorer}\label{sec:emission}
	As shown in Fig. \ref{fig:framework}, the emission scorer takes the pair-wise embeddings
	as input and independently assigns each token an emission score with regard to each label.
	%As mentioned above, every domain has a domain-specific label set $L_D = \left\{\ell_1, \ell_2, \cdots, \ell_{t_D} \right\}$. 
	%The Emission Scorer component assigns a token an emission score with regard to each label $\ell_j$. 
	This component can be implemented with any few-shot classification model. 
	% We employ matching network \citep{matching} 
	% and prototypical network \citep{prototypical} first to show the generalization capability of our framework. \atmasay{Change here if change experiment style.}
	% Then, we propose the normalized match network (NMN) which is a normalized version of matching network.  
	We employ 3 different models: 
	matching network \citep{matching}, prototypical network \citep{prototypical} 
	and normalized match network (NMN) which is our normalized version of matching network. 
	
	For matching network, the emission score is calculated as:
	\begin{equation} \nonumber
	f_E^M(y_j, j, \bm{x}, \mathcal{S}) = \sum_{k=1}^{N_{W}} \mathbb I(\ell_k = y_j) \cdot \mathrm{Sim}(e_j, e_k^S)
	\end{equation}
	$N_{W}$ is the number of words in support set $S$, 
	and $\ell_k$ is the label of the $k_{th}$ word in $S$. $ \mathbb I(b)$ is the indicator function where $\mathbb I(b)=1$ if $b=True$ (otherwise $\mathbb I(b)=0$). 
	We denote $e_i$ and $e_k^S$ as the embedding of $x_i$ and $k_{th}$ word in $S$, respectively. 
	Then $\mathrm{Sim}(e_j, e_k^S)$ is a function that calculates the similarity between them. 
	We utilize dot product as the similarity metric for all the models, 
	but it is also possible to use other similarity metrics. 
	%In our case, $\mathrm{Sim}(e_j, e_k^S) = e_j \cdot e_k^S$. 
	In our case, $Sim(e_j, e_k^S)$ could be calculated as:
	%\begin{equation} \nonumber
	%Sim(e_j, e_k^S) = e_j \cdot e_k^S
	%\end{equation}
	$Sim(e_j, e_k^S) = e_j \cdot e_k^S$

	For normalized matching network, we normalize the emission score for each label as follows:
	\begin{equation} \nonumber
	f_E^S(y_j, j, \bm{x}, \mathcal{S}) = \frac{1}{M_{y_j}} \cdot \sum_{k=1}^{N_{W}} \mathbb I(\ell_k = y_j) \cdot \mathrm{Sim}(e_j, e_k^S) 
	\end{equation}
	where $M_{y_j} = \sum_{k=1}^{N_{W}} \mathbb I(\ell_k = y_j)$. 
	%\[
	%M_{y_j} = \sum_{k=1}^{N_{W}} \mathbb I(\ell_k = y_j)
	%\]
	
	For prototypical network, we first calculate an $h$ dimensional prototype representation $c_{\ell_m}$ for each label:
	$
	c_{\ell_m} = \frac{1}{M_{\ell_m}}\sum_{k=1}^{N_{W}} \mathbb I(\ell_k = \ell_m) \cdot e_k^S 
	$
	,where $M_{\ell_m} = \sum_{k=1}^{N_{W}} \mathbb I(\ell_k = \ell_m)$. 
	%\[
	%M_{\ell_m} = \sum_{k=1}^{N_{W}} \mathbb I(\ell_k = \ell_m)
	%\]
	Then the emission score is calculated as: 
	\begin{equation} \nonumber
	f_E^P(y_j, j, \bm{x}, \mathcal{S}) = \mathrm{Sim}(e_j, c_{y_j}) 
	\end{equation}

	\begin{table}[t]
		\centering
		\footnotesize
		\begin{tabular}{SlllSS} \toprule
			{\textbf{Task}}  & {\textbf{Dataset}}  & {\textbf{Domain}}  & {\textbf{\# Sent}} & {\textbf{\# Labels}}  \\ 
			\midrule
			{\multirow{7}{*}{\makecell{Slot\\Tagging}}}	& {\multirow{7}{*}{Snips}}	& {\small{Weather}}& {2,100}  & {10}   \\		
			& & {\small{Music}}& {2,100}  & {10}  \\
			& & {\small{PlayList}} & {2,042} & {6}  \\		
			& & {\small{Book}}& {2056}  & {8} \\
			& & {\small{SearchScreen}} & {2,059}  & {8}  \\
			& & {\small{Restaurant}}& {2,073}  & {15}   \\		
			& & {\small{CreativeWork}}& {2,054}  & {3}  \\
			\midrule
			{\multirow{4}{*}{NER}} & {\small{CoNLL}}	& {\small{News}}  & {20679}  & {5}   \\
			& {\small{GUM}}	& {\small{WiKi}} & {3,493}  & {12}    \\
			& {\small{WNUT}} & {\small{Social}} & {5,657} & {7}    \\
			& {\small{OntoNotes}}  & {\small{Mixed}}      & {159,615}   & {19}       \\
			\bottomrule
		\end{tabular}
		\caption{\footnotesize
			Statistic of Original Dataset}\label{tbl:dataset_old}
		\vspace*{-3mm}
	\end{table}
	
	\begin{table*}[t]
		\centering
		\footnotesize
		%		\caption{Bi-gram accuracy on slot tagging task. Bi-grams are classified according to label types of B, I, O.}
		\begin{tabular}{cccccccccccc}
			\toprule
			\multirow{2}{*}{\textbf{Domain}} 					&
			\multicolumn{7}{c}{\textbf{Slot Tagging}} &
			\multicolumn{4}{c}{\textbf{Named Entity Recognition}} 	\\
			\cmidrule(lr){2-8}
			\cmidrule(lr){9-12}
			& {\textbf{We}} & {\textbf{Mu}} & {\textbf{Pl}} & {\textbf{Bo}} & {\textbf{Se}} & {\textbf{Re}} & {\textbf{Cr}} & {\textbf{News}} & {\textbf{Wiki}} & {\textbf{Social}} & {\textbf{Mixed}}\\	
			\midrule
			{\textbf{Ave. $\bm{|S|}$}} (1-shot)  & 6.15 & 7.66 & 2.96 & 4.34 & 4.29 & 9.41 & 1.30 & 3.38 & 6.50 & 5.48 & 14.38 \\
			{\textbf{Samples}} (1-shot)   & 2,000 & 2,000 & 2,000 & 2,000 & 2,000 & 2,000 & 2,000 & 4,000 & 4,000 & 4,000 & 4,000  \\ 
			%		\cmidrule(lr){2-8}
			%		\cmidrule(lr){9-12}
			\midrule
			{\textbf{Ave. $\bm{|S|}$}} (5-shot)  & 28.91 & 34.43 & 13.84 & 19.83 & 19.27 &  41.58 & 5.28 & 15.58 & 27.81 & 28.66 & 62.28 \\
			{\textbf{Samples}} (5-shot)   & 1,000 & 1,000 & 1,000 & 1,000 & 1,000 & 1,000 & 1,000 & 1,000 & 1,000 & 1,000 & 1,000  \\
			\bottomrule
		\end{tabular}
		\caption{\footnotesize{Overview of few-shot data. Here, ``Ave. $|S|$" corresponds to the average support set size of each domain. And ``Sample" stands for the number of few-shot samples we build from each domain.}
		}\label{tbl:new_dataset}
		\vspace*{-3mm}
	\end{table*}

	%We examine the proposed framework on two NLP sequence labeling tasks: 
	%Name Entity Recognition(NER) and Slot Tagging(ST). 
	%During both training and testing phrases, all the models are asked to label a sample sentence singly based on its corresponding support set. 
	%We explore the $1$-$shot$ sequence labeling case in all the experiments.
	\section{Experiment}\label{sec:exp}
	We exploit multiple datasets to evaluate the proposed methods on two learning tasks: {\it name entity recognition} (NER) and {\it slot tagging}. 
	We present the detailed results under the 1-shot and 5-shot setting, 
	which transfers the learned knowledge from source domains (training) to an unseen target domain (testing) containing only a 1-shot/5-shot support set.

	%Table \ref{tbl:dataset} is a summarization of the datasets used in our experiments. 
	%To simulate the few-shot scenario, 
	%we construct few-shot data from the existing datasets \citep{DBLP:journals/corr/abs-1805-10190, sang2003introduction, pradhan2013towards, zeldes2017gum, derczynski2017results}. 
	%We pair each $1$-$shot$ support set $S$ with an $(\bm{x},\bm{y})$ pair that does not belong to it, 
	%and call the combination of them as a $1$-$shot$ sample.
	%For ST task, we independently sample 100 $1$-$shot$ support sets from each domain with Algorithm \ref{algorithm}. Then for each support set, we built 20 $1$-$shot$ samples by respectively pairing it with 20 different $(\bm{x},\bm{y})$ pairs. In total, we obtain 2000 $1$-$shot$ samples for each domain. Similarly, for NER task, we obtain 4000 $1$-$shot$ samples for each domain by sampling 200 $1$-$shot$ support sets together with 20 different $(\bm{x},\bm{y})$ pairs for each support set.
	
	\subsection{Settings}
	%\subsection{Dataset}
	\noindent\textbf{Dataset}
	%We examine the proposed framework on two NLP sequence labeling tasks: 
	%Name Entity Recognition(NER) and Slot Tagging(ST). 
	%During both training and testing phases, all the models are asked to label a sample sentence singly based on its corresponding support set. 
	%We explore the $1$-$shot$ sequence labeling case in all the experiments.
	%\paragraph{Data source} 
	For slot tagging, we exploit the \texttt{snips} dataset \citep{DBLP:journals/corr/abs-1805-10190}, 
	because it contains 7 domains with different label sets and is easy to simulate the few-shot situation.
	%The \texttt{snips} dataset includes data from 7 domains, which are 
	The domains are 
	Weather (We), Music (Mu), PlayList (Pl), Book (Bo), Search Screen (Se), Restaurant (Re) and Creative Work (Cr).
	For named entity recognition, we utilize 4 different datasets: 
	\texttt{CoNLL-2003} \citep{sang2003introduction}, \texttt{GUM} \citep{zeldes2017gum}, \texttt{WNUT-2017} \citep{derczynski2017results} and \texttt{Ontonotes} \citep{pradhan2013towards}, 
	each of which contains data from only 1 domain. 
	The 4 domains are News, Wiki, Social and Mixed. 
	Information of original datasets is shown in Table \ref{tbl:dataset_old}.
	
	To simulate the few-shot situation, 
	we construct the few-shot datasets from original datasets, 
	where each sample is the combination of a query data $(\bm{x^q},\bm{y^q})$ and corresponding k-shot support set $\mathcal{S}$.
	Table \ref{tbl:new_dataset} shows the overview of experiment data we utilize.
	
	\noindent\textbf{Few-shot Data Construction}
	%A big difference between few-shot and conventional sequence labeling is the definition of the sample. 
	%In few-shot learning, a sample is no longer an $(\bm{x^q},\bm{y^q})$ pair. 
	%Instead, it's the combination of an $(\bm{x^q},\bm{y^q})$ pair and a corresponding k-shot support set.
	Here, we take the 1-shot setting as an example to illustrate the data construction procedure.
	Specifically, for each domain in slot tagging, 
	we randomly sample 2,000 query data and 100 different 1-shot support sets. 
	Then, we construct 2,000 samples by letting every 20 query data attach to 1 same support set.
	Similarly, for 1-shot NER task, 
	we build 4,000 samples with 200 different k-shot support sets for each domain. 
	Each k-shot support set here is obtained with the Algorithm \ref{algorithm}.\footnote{
		In practice, our algorithm is likely to exclude some extreme samples situation. 
		For example, it may include only co-occurrence cases and exclude all along cases for some label pairs. 
		So, we randomly keep 20\% of samples that should have been deleted to make the sampling more uniform.
	}
	
	\noindent\textbf{Evaluation}
	To test the robustness of our framework, 
	we perform cross validation by separately testing models on different domains.
	Each time,
	we pick one target domain for testing, one domain for development, 
	and use the rest domains as source domains for training. 
	So for slot tagging, 
	all models are trained on 10,000 samples, and validated as well as tested on 2,000 samples respectively.
	And for named entity recognition, 
	we train the models on 8,000 samples, and then validate and test them on 4,000 samples.
	%Figure \ref{fig:overview} shows the procedure of a single cross validation.
	
	During evaluation phase, we aggregate samples with the same support set together 
	% and call the combination of these samples 
	as one testing episode.
	We take the average of the macro F1 scores over all the episodes as the evaluation metric. 
	For each episode, 
	Macro F1 scores are calculated as 
	$
	F = \frac{2 P R}{P + R}
	$
	, where $P = \sum_{k=1}^{K} \frac{P_k}{K}$ and $R = \sum_{k=1}^{K} \frac{R_k}{K}$. 
	Here, $P_k$ and $R_k$ are the precision and recall of the $k_{th}$ sample of an episode, 
	and $K$ is the number of samples in an episode.\footnote{
		We calculate F1 score with \texttt{conlleval} script:  \url{https://github.com/sighsmile/conlleval}}
	
	%\paragraph{Hyperparameters}
	\noindent\textbf{Hyperparameters}
	To control the effect of nondeterministic of neural network training \citep{reimers-gurevych:2017:EMNLP2017}, 
	we run each experiment with three different random seeds and report the average scores of them.
	For all the models built under our framework,
	we use the uncased \texttt{BERT-Base} \citep{BERT}
	to calculate contextual embedding, 
	and all models are trained with ADAM algorithm \citep{DBLP:journals/corr/KingmaB14} with batch size 4 and a learning rate of 1e-5. 
	For all models that take the dependency transfer mechanism (DT) into account, 
	we implement a learnable scaling parameter initialized randomly in a range of 0 to 1.
	Early stop in training is performed when there is no loss decay for 2 epochs.

	\begin{table*}[t]
		\centering
		\footnotesize
		\begin{tabular}{p{2cm}SSSSSSSSSSSSSS}\toprule
			\multirow{2}{*}{\textbf{Model}} &
			\multicolumn{7}{c}{\textbf{Slot Tagging}} & &
			\multicolumn{4}{c}{\textbf{Named Entity Recognition}}  &	\\
			\cmidrule(lr){2-8}
			\cmidrule(lr){10-13}
			
			& {\textbf{We}} & {\textbf{Mu}} & {\textbf{Pl}} & {\textbf{Bo}} & {\textbf{Se}} & {\textbf{Re}} & {\textbf{Cr}} & {\textbf{Ave.}} 
			& {\textbf{News}} & {\textbf{Wiki}} & {\textbf{Social}} & {\textbf{Mixed}} & {\textbf{Ave.}} \\ 
			\midrule
			%		\cmidrule(lr){2-8}
			%		\cmidrule(lr){10-13}
			{Bi-LSTM}  & 10.54& 16.93& 17.75& 54.01& 17.48& 23.10& 9.7 5&  21.37 
			& 2.60& 3.07& 0.56& 1.91& 2.04	\\
			{SimBERT }       & 36.10&  37.08& 35.11& 68.09& 41.61& 42.82& 23.91& 40.67
			& 19.22& \ \ {\textbf{6.91}} & 5.18& 13.99&  11.35 \\
			{TransferBERT }  & 45.66& 27.80& 39.27& 9.28& 13.59& 35.12& 9.97& 25.81 
			& 1.62& 0.58& 1.38& 3.08& 1.67 \\
			{WPZ}			& 4.34& 7.12& 13.51& 40.10& 11.85& 8.02& 9.36& 13.47 
			& 3.63 & 2.00 & 0.92 & 0.67 &  1.80 \\
			{WPZ+GloVe} & 17.95 & 22.08 & 19.90 & 42.67 & 22.28 & 22.74 & 16.86 & 23.50 
			& 9.36 & 3.24 & 2.30 & 2.55 &  4.36 \\ 
			% pair-wise result
			%		{WPZ+BERT}  & 61.94& 42.34& 52.54& 70.49& 65.58& 57.64& 64.59& 59.30
			%		& 33.00& 3.50& 10.37& 6.84&13.43 \\ 
			%		{MN}   & 50.85 & 31.99& 42.83& 67.37& 37.96 & 43.17& 54.78& 46.99 
			%		& 10.42& 4.16 & 14.23 & 2.86 & 7.92 \\ 
			% sep result
			{WPZ+Bert} & 47.54 & 39.77 & 50.01 & 69.69 & 60.05 & 54.32 & 67.14 & 55.51 
			& 31.85 & 3.69 & 9.16 & 6.15 & 12.71 \\
			{MN}  & 20.85 & 10.91 & 40.22 & 58.12 & 24.48 & 32.98 & {\textbf{70.17}} & 36.82 
			& 19.44 & 4.93 & 13.39 & 14.04 & 12.95\\
			%		\cmidrule(lr){2-8}
			%		\cmidrule(lr){10-13}	
			\midrule
			% pair-wise  result
			%		{NMN}           & 63.17 & 46.04 & 51.72 & 72.50 & 65.20 & 50.92 & { \textbf{68.50}} & 59.72
			%		& 31.05& 4.38& 10.19& 7.28&  13.23 \\
			% sep version result
			{NMN} & 46.78 & 39.99 & 53.53 & 68.54 & 58.65 & 57.64 & 67.25 & 56.05 
			& 32.27 & 3.88 & 10.43 & 6.59 & 13.29 \\
			{MN+P\&D}     & 56.27&  {\textbf{53.30}}& 59.14& 75.88& 54.45& 55.33& 41.10& 56.50
			& 7.42& 4.45&  {\textbf{24.08}}& 8.89& 11.21 \\
			{WPZ+Bert+P\&D}      & {\textbf{69.36}}& 52.51& 58.75& 79.29& {\textbf{71.36}}&  66.03& 59.39& 65.24
			& 47.49& 3.17& 21.06& 25.28&24.25 \\ 
			{NMN+P\&D}   & 68.68& 53.12&  {\textbf{67.39}}&  {\textbf{80.92}}&  71.25& {\textbf{69.32}}& 60.26 &  {\textbf{67.27}}
			& {\textbf{47.94}}& 3.48& 23.01& {\textbf{26.10}}&   {\textbf{25.13}} \\ 
			\bottomrule
		\end{tabular}
		\caption{ \footnotesize  
			F1 scores on 1-shot slot tagging and name entity recognition. 
			%		We respectively explore WarmProtoZero (WPZ), matching network (MN), normalized matching network (NMN) and the dependency transfer mechanism (DT) in our experiments. 
			\texttt{+P\&D} denotes that model is equipped with both pair-wise embedding and dependence transfer. 
			Score below mid-line are from our models, which achieve the best performance. 
			Ave. shows the averaged scores. 
		}\label{tbl:1shot}
	\end{table*}
	
	\begin{table*}[t]
		\centering
		\footnotesize
		\begin{tabular}{p{2cm}SSSSSSSSSSSSSS}\toprule
			\multirow{2}{*}{\textbf{Model}} &
			\multicolumn{7}{c}{\textbf{Slot Tagging}} & &
			\multicolumn{4}{c}{\textbf{Named Entity Recognition}}  & 	\\
			\cmidrule(lr){2-8}
			\cmidrule(lr){10-13}
			& {\textbf{We}} & {\textbf{Mu}} & {\textbf{Pl}} & {\textbf{Bo}} & {\textbf{Se}} & {\textbf{Re}} & {\textbf{Cr}} & {\textbf{Ave.}} 
			& {\textbf{News}} & {\textbf{Wiki}} & {\textbf{Social}} & {\textbf{Mixed}} & {\textbf{Ave.}} \\ \midrule
			{Bi-LSTM}  & 25.44 & 39.69 & 45.36 & 73.58 & 55.03 & 40.30 & 40.49 & 45.70  
			& 6.59 & 8.34 & 0.87 & 12.20 & 7.00 \\
			{SimBERT }       & 53.46 & 54.13 & 42.81 & 75.54 & 57.10 & 55.30 & 32.38 & 52.96 
			& 32.01 & 10.63 & 8.20 & 21.14 & 18.00\\
			{TransferBERT }    & 56.01 & 43.85 & 50.65 & 14.19 & 23.89 & 36.99 & 14.29 & 34.27 
			& 4.93 & 0.91 & 3.71 & 15.64 & 6.30 \\
			{WPZ}     & 9.35 & 14.04 & 16.71 & 47.23 & 19.56 & 11.45 & 13.41 & 18.82 
			& 4.15 & 3.13 & 0.89 & 0.90 & 2.27 \\ 
			{WPZ+GloVe}     & 27.15 & 34.09 & 22.15 & 50.40 & 28.58 & 34.59 & 23.76 & 31.53 
			& 16.84 & 5.26 & 5.42 & 3.51 & 7.76 \\ 
			% pair-wise  result
			%		{WPZ+BERT}    & 66.80 & 57.20 & 47.11 & 75.06 & 75.43 & 65.27 & {\textbf{70.30}} & 65.31 
			%		& 46.41 & 8.34 & 17.29 & 13.17 & 21.30 \\ 		
			%		{MN}     & 42.94 & 49.37 & 54.80 & 65.80 & 39.74 & 29.68 & 55.47 & 48.26 
			%		& 21.25 & 3.96 & 8.99 & 8.29 & 10.62 \\ 
			% sep result
			{WPZ+Bert} & 69.06 & 57.97 & 44.44 & 71.97 & 74.62 & 51.01 & 69.22 & 62.61 
			& 49.85 & 9.60 & 18.66 & 13.17 & 22.82 \\
			{MN}  & 38.80 & 37.98 & 51.97 & 70.61 & 37.24 & 34.29 & {\textbf{72.34}} & 49.03
			& 20.33 & 5.73 & 7.84 & 8.04 & 10.49 \\
			\midrule
			% pair-wise  result
			%		{NMN}             & 66.46 & 58.13 & 47.65 & 78.21 & {\textbf{80.29}} & 67.35 & 69.90 & 66.86 
			%		& 46.50 & 9.22 & 18.97 & 14.09 & 22.20 \\
			% sep result
			{NMN}  & 63.76 & 55.38 & 43.81 & 73.72 & 73.37 & 64.71 & 71.69 & 63.78
			& 49.94 & 9.13 & 16.55 & 11.10 & 21.68 \\
			{MN+P\&D}     & 51.30 & 61.00 & 68.73 & 70.93 & 59.84 & 33.03 & 47.46& 56.04 
			& 17.53 & 4.80 & 6.68 & 12.29 & 10.33 \\ 
			{WPZ+Bert+P\&D}        & {\textbf{71.38}} & {\textbf{67.48}} & 71.38 & 69.33 & 58.09 & 72.81 & 68.05 & 68.36 
			& {\textbf{49.25}} & 10.85 & 31.81 & {\textbf{26.98}} & {\textbf{29.72}} \\ 
			{NMN+P\&D}     & 70.90 & 64.63 & {\textbf{75.19}} & {\textbf{83.64}} & {\textbf{78.23}} & {\textbf{78.07}} & 62.73 & {\textbf{73.27}} 
			& 46.56 & {\textbf{12.46}} & {\textbf{33.59}} & 22.81 & 28.86 \\ 
			\bottomrule
		\end{tabular}
		\caption{ \footnotesize.  
			F1 score results on 5-shot slot tagging and and name entity recognition. 
			% 		\texttt{+P\&D} denotes that model is equipped with both pair-wise embedding and dependence transfer. 
			% 		Score below mid-line are from our models, which achieve the best performance.
			Our methods achieve the best performance.
		}\label{tbl:5shot}
		\vspace*{-4mm}
	\end{table*}

	\subsection{Baselines}
	We compare our model with the following baselines:
	
	\textbf{Bi-LSTM} is a bidirectional LSTM sequence labeling model \citep{huang2015bidirectional}. 
	For each sample, it is trained on the support set and tested on the query sample. 
	%corresponding $(\bm{x},\bm{y})$ pair.
	%We use GloVe \citep{GloVe} as token embedding. 
	
	\textbf{SimBERT} is a model that predicts labels according to cosine similarity between contextual token embeddings given by a non-fine-tuned BERT model. 
	For each token $x_j$, SimBERT finds it's most similar token $x_k'$ in support set, and the label of $x_j$ is predicted to be the label of $x_k'$. 
	
	\textbf{TransferBERT} is a domain transfer model based on BERT. 
	It performs sequence labeling following the NER setting of BERT \citep{BERT}. 
	We first fine-turn the pre-trained BERT model on the sources domains
	and then further fine-tune it on the support set of the target domain. 
	%As transfer learning is the mainstream approach for the few-shot scenario, this is a very strong baseline. 
	
	\textbf{WarmProtoZero (WPZ)} \citep{baseline} is a few-shot sequence labeling model 
	that regards sequence labeling as classification of each single token. 
	It pre-trains a prototypical network \citep{prototypical} on source domains, 
	and utilize it to do token-level classification on target domains without training. 
	\citet{baseline} use randomly initialized token embeddings.
	To eliminate the influence of different embedding methods, 
	we further implement WPZ with pre-trained \textbf{GloVe} embedding \citep{GloVe} and contextual embedding given by \textbf{BERT}.  
	%\textbf{WPZ+BERT} is an implementation of WPZ model under our framework. 
	%To eliminate the influence of different embedding methods, we implement it
	%with Contextual Embedding and Emission Scorer components of our framework. Then we assign each token a label singly based on the emission score. 
	
	\textbf{Matching Network (MN)} is similar to WPZ. 
	The only difference is that we employ matching network \cite{matching} with \textbf{BERT} embedding for classification.

	\subsection{Main Results}
	\noindent\textbf{Results of 1-shot Setting}
	Table \ref{tbl:1shot} shows the 1-shot sequence labeling results on slot tagging and NER tasks. 
	Each column respectively shows the F1 scores of taking a certain domain as target domain (test).
	As shown in the tables, 
	our normalized matching network equipped with the pair-wise embedding and dependency transfer mechanism (P\&D) 
	achieves the best performance in both ST and NER tasks. 
	It outperform the strongest few-shot learning baseline by average F1 scores of 11.76 and 12.18 on two tasks.
	More importantly, 
	all models boosted by the pair-wise embedding and dependency transfer (P\&D) 
	gain a significant increasing on F1 score in all domains. 
	This reflects the powerful adaptability and effectiveness of our proposed methods. 
	
	Our model significantly outperforms Bi-LSTM and TransferBERT, 
	indicating that the number of labeled data under few-shot setting is too scarce for 
	both conventional machine learning and transfer learning models. 
	% to achieve a satisfied performance.
	Moreover, the performance of SimBERT demonstrates the superiority of metric-based methods 
	over conventional machine learning models in few-shot setting. 
	
%	Although the original WarmProtoZero (WPZ) model is constructed upon the metric-based method,
%	it suffers from the weak representation ability of its token embeddings. 
	%which are learned during training phase.
	%We therefore employ GloVe to initialize its embedding, and further implement it under our framework which employs BERT to calculate contextual embedding.
	The original WarmProtoZero (WPZ) model suffers from the weak representation ability of its token embeddings. 
	After implemented with pre-trained GloVe embedding and contextual embedding from BERT model, 
	it's performance improves significantly.
	This shows the importance of embedding in few-shot setting. 
	%Comparing to other baseline models, 
	%WPZ+BERT and MN achieve much better performance. 
	%Such results demonstrate the effectiveness and generalization capability of our framework.
	%More importantly, although been implemented under the same framework, our model still significantly outperforms them.
	%We attribute the improvement to the dependency transfer mechanism and the normalized matching network, and we respectively discuss the effectiveness of them in the next section.
	
	\noindent\textbf{Results of 5-shot Setting}
	To verify the proposed model's generalization ability in more shots situations, 
	we perform 5-shots experiments for the two tasks. 
	The results are shown in Table \ref{tbl:5shot}, which is consistent with 1-shot setting in general trending. 
	Our model achieves the best performance on average F1 score. 
	Dependency transferring and pair-wise embedding together consistently improve models performance.

	\subsection{Analysis}
	\noindent\textbf{Ablation Test}
	To get further a understanding of each component in our method (NMN+P\&D), 
	we conduct ablation analysis on slot tagging and NER tasks of both 1-shot and 5-shot setting.
	Each component of our method is removed respectively, 
	including: 
	\texttt{dependency transferring}, \texttt{pair-wise embedding}, \texttt{learnable scaling}, \texttt{label-wise normalization}.
	%dependence transferring, pair-wise embedding, learnable scaler, normalize emission.
	
	When dependency transferring is removed, 
	we directly predict labels with emission score and huge F1 score drops are witnessed on all settings. 
	This ablation demonstrates a great necessity for considering label dependency. 
	
	For our method without pair-wise embedding, 
	we represent query and support sentence independently, which lead to significant drops in all settings. 
	We address the drop to the fact that support and query sentences always contains similar structure and context, 
	and pair-wise embedding can help remove uncertainty on token representations. 
	
	If we replace the learnable scaling $\lambda$ with a hand-tuned hyperparameter, 
	the model appears to be poorly trained and thus its performance drops a lot.
	This is because the automatic scaling can smooth the training process by balancing the value scale of emission and transition scores
	and assigning proper importance weight to the two scores. 
	
	And when we remove label-wise normalizing, 
	the emission scorer regress to matching network and predicts labels poorly. 
	This indicates that modeling the label frequency latently will confuse the model and label-wise normalization can ease the problem.

	\begin{table}[t]
		\centering
		\footnotesize
		%	\begin{tabular}{lSSSS}
		\begin{tabular}{lcccc}
			\toprule
			\multirow{2}{*}{\textbf{Model}} & \multicolumn{2}{c}{\textbf{Slot Tagging}} & \multicolumn{2}{c}{\textbf{NER}} \\
			\cmidrule(lr){2-3}
			\cmidrule(lr){4-5}
			& {1-shot} & {5-shot} & {1-shot} & {5-shot} \\
			\midrule
			Ours & 67.27 & 73.27 & 25.13 & 28.86  \\
			~ - dependency transfer & -8.90 & -6.41 & -11.84 & -9.02  \\
			~ - pair-wise embedding & -5.57 & -5.33 & -8.48 & -2.86  \\
			%		~ - train transition & -4.50 & -0.85 & -10.39 & -4.22  \\
			~ - learnable scaling & -26.61 & -29.70 & -16.28 & -18.78  \\
			~ - label-wise normalize & -10.97 & -17.23 & -13.92 & -18.62  \\
			\bottomrule
		\end{tabular}
		\caption{
			\footnotesize
			Ablation test over different components on slot tagging and NER task. 
			% 		Each time we remove one component and report the averaged F1-score of all domains.
			Results are averaged F1-score of all domains.
		}\label{tbl:ablation}
	\end{table}

	\begin{table}[t]
		\centering
		\footnotesize
		%	\begin{tabular}{lSSSS}
		\begin{tabular}{lcccc}
			\toprule
			\multirow{2}{*}{\textbf{Model}} & \multicolumn{2}{c}{\textbf{Slot Tagging}} & \multicolumn{2}{c}{\textbf{NER}} \\
			\cmidrule(lr){2-3}
			\cmidrule(lr){4-5}
			& {1-shot} & {5-shot} & {1-shot} & {5-shot} \\
			\midrule
			NMN+PW & 58.37 & 66.86 & 13.29 & 19.84  \\
			NMN+PW+Rule & 62.77 & 72.42 & 14.74 & 24.64  \\
			NMN+PW+DT & \textbf{67.27} & \textbf{73.27} & \textbf{25.13} & \textbf{28.86}  \\
			\bottomrule
		\end{tabular}
		\caption{
			\footnotesize
			Comparison between transition rules and dependency transfer. 
			Results are averaged F1-score over all domains. 
			\texttt{PW} is pair-wise embedding and \texttt{DT} denotes dependency transfer.
		}\label{tbl:rule}
		\vspace*{-3mm}
	\end{table}
	
	% horizontal version
	%\begin{table*}[t]
	%	\centering
	%	\footnotesize
	%	\label{tbl:error_rd}
	%	\begin{tabular}{cccccccc}
	%		\toprule
	%		\multicolumn{1}{c}{\textbf{Bi-gram Type}} 					&
	%		\multicolumn{4}{c}{\textbf{Tag Border}} &
	%		\multicolumn{2}{c}{\textbf{Tag Consistency}} &
	%		\multicolumn{1}{c}{\textbf{Other}}		\\
	%		\cmidrule(lr){2-5}
	%		\cmidrule(r){6-7}
	%		&
	%		\multicolumn{1}{c}{\textbf{O-O}} &
	%		\multicolumn{1}{c}{\textbf{O-B}} &
	%		\multicolumn{1}{c}{\textbf{B-O}} &
	%		\multicolumn{1}{c}{\textbf{I-O}} &
	%		\multicolumn{1}{c}{\textbf{B-I}} &
	%		\multicolumn{1}{c}{\textbf{I-I}} & \\
	%		\midrule
	%		\textbf{Proportion} & 25.32\% & 21.75\% & 7.28\% & 5.12\% & 11.84\% & 10.79\% & 17.89\% \\
	%		\textbf{NMN} & 80\% & 71\% & 73\% & 83\% & 71\% & 72\% & 89\% \\
	%		\textbf{NMN+DT} & 89\% & 71\% & 88\% &  92\% & 76\% & 77\% & 87\% \\
	%		\bottomrule
	%		
	%	\end{tabular}
	%	\caption{\footnotesize{
	%			Accuracy analysis of label prediction on 1-shot slot tagging. 
	%			According to label types, bi-grams are classified into categories of: O-O, O-B, B-O, I-O, B-I, I-I and Other.
	%			Results present models' accuracy over different types of bi-gram. 
	%			Proportion denotes the proportion of different bi-gram types among dataset. }
	%	}
	%\end{table*}

	% vertical version
	\begin{table}[t]
		\centering
		\footnotesize
		\begin{tabular}{llScc}
			\toprule
			\multicolumn{2}{c}{\textbf{Bi-gram Type}} & \textbf{Proportion} & \textbf{NMN+PW} & \textbf{+DT} \\
			\cmidrule(lr){1-2}
			\cmidrule(lr){3-5}
			\multirow{4}{*}{\makecell{Border}} 
			& O-O  & 25.32\%  & 80\% & \textbf{89\%} \\ 
			& O-B  & 21.75\%  & 71\% & \textbf{71\%} \\
			& B-O  & 7.28\%   & 73\% & \textbf{88\%} \\
			& I-O  & 5.12\%   & 83\% &  \textbf{92\%} \\
			& B-I/I-B & 6.76\% & \textbf{77\%} & 75\% \\
			\midrule
			\multirow{2}{*}{\makecell{Inner}} 
			& B-I  & 11.84\%  & 71\% & \textbf{76\%} \\
			& I-I  & 10.79\%  & 72\% & \textbf{77\%} \\
			\midrule
			Start
			& S-B  & 0.23\%   & 69\% & \textbf{72\%} \\
			& S-O  & 10.90\%  & \textbf{97\%} & 95\% \\
			\bottomrule
		\end{tabular}
		\caption{\footnotesize
			Accuracy analysis of label prediction on 1-shot slot tagging. 
			%			According to label types, bi-grams are classified into categories of: O-O, O-B, B-O, I-O, B-I, I-I and Other.
			%			Results present models' accuracy over different types of bi-gram. 
			The table shows accuracy and proportion of different bi-gram types in dataset.
			\texttt{S} denotes \texttt{START} label.
		}\label{tbl:error_rd}
		\vspace*{-3mm}
	\end{table}

	\noindent\textbf{Effectiveness of Dependency Transfer}
	While dependency transfer brings significant improvements, 
	two natural questions arise:  whether dependency transferring just learns simple transition rules and why it works. 
	
	To answer the first question, 
	we compare dependency transferring with transition rules in Table \ref{tbl:rule}.\footnote{
		Rule:
		We greedly predict the label for each token and block the result that conflicts with previous label. 
%		for each token, we block prediction of the labels that is illegal according to previous label. 
%		For example, we block the prediction of $I-l_1$ after $B-l_2$, where $l_1$
	} 
	Results show that transition rules help to correct prediction on all tasks and settings.
	However, label dependency transferring can further improve performance.
	
	To have a deeper insight of the effectiveness of DT, 
	we conduct accuracy analysis for dependency transfer. 
	%of label prediction for NMN and NMN+DT under 1-shot slot tagging. 
	We assess the label predicting accuracy of different types of label bi-grams. 
	The result is shown in Table \ref{tbl:error_rd}. 
	We futher summarize the bi-grams into 3 categories: \texttt{Border}, \texttt{Inner} and \texttt{Start}.
	\texttt{Border} includes the bi-grams across the border of a tag span; 
	\texttt{Inner} is the bi-grams within a tag span.
	% As showed in table,  
	Improvements of \texttt{Inner} show that dependency transfer reduces illegal label transition successfully. 
	%For example, 
	%NMN may predict wrong consequent labels of B-time and I-city, 
	%which is very likely to be fixed by the dependency transfer mechanism. 
	Interestingly, 
	results of \texttt{Border} show the dependency transfer mechanism also helps to decide the boundaries of label spans more accurately, 
	which is hard to achieve by adding transition rules.

	\section{Related Works}	
	Here, we introduce related works of sequence labeling
	%(Section \ref{sec:rw_seq_l}) 
	and few-shot learning.
	
	\noindent\textbf{Sequence Labeling} \label{sec:rw_seq_l}
	Conditional random fields 
	are
%	have been shown to be 
	one of the most successful approaches for sequence labeling \citep{huang2015bidirectional,ma2016end,CRF}. 
	It finds the optimal label sequence by considering both emission 
	and label transition score.
	Recent models take advantage of neural models to calculate emission score, 
	such as CNN \citep{cnn}) 
	and bi-LSTM \citep{lstm}.
	%And label dependency is modeled as transition probability between labels (transition score). 
	However, these models suffer when the data is scarce. 
	
	For sequence labeling under data scarcity situation, 
	transfer learning and few-shot learning have been introduced. 
	%\citet{yang2017transfer} constructs a deep hierarchical recurrent neural network to transfer knowledge from source domain to the target domain by sharing the hidden feature representation and part of the model parameters between them.
	\citet{yang2017transfer} propose to transfer knowledge of hidden feature representation and part of the model parameters.
	\citet{jha2018bag} develop a Bags-of-Architecture for re-using source domain model. 
	%The source domain model can be viewed as a common feature extractor.
	%Different from us, 
	%transfer models are trained on target domain and focus on leveraging the label schema relations between different domains. 
	%More important, the number of training data used by these model is still much more than our few-shot case.
	\citet{baseline} explore few-shot named entity recognition 
	by independently classifying each token with prototypical network \citep{prototypical}.
	%However, training a model from scratch for every new domain is time-consuming, and the model is very  likely to overfit when there is little training data \citep{dropout}.

	\noindent\textbf{Few-shot Learning} \label{sec:rw_fsl}
	%Human always adapt to those new tasks quickly without needing too many explanations or examples due to their prior experience. 
	%But traditional machine learning models usually not built upon an assumption that other knowledge is potentially helpful. 
	Few-shot learning is built upon the assumption that prior knowledge is potentially helpful. 
	%Early works on few-shot learning focused primarily on computer vision tasks, 
	%especially those in object classification. 
	%Some researchers present to classify an sample by calculating its distances between different samples using hand-crafted feature \citep{fink2005object}. 
	Traditional methods depend highly on hand-crafted features \cite{fei2006knowledge,fink2005object}. 
	% Transferring knowledge from object classes previously learned was also introduced \citep{fei2006knowledge}. 
	% However, these methods suffer from poorly generalization capability. 
	Recent efforts primarily focus on metric learning \cite{prototypical,matching}.
	%With the development of a variety of deep learning methods, 
	%some models with great generalization and feature representation capabilities have been introduced. 
	%\citet{prototypical} proposes to learn a 
	%central representation for every class and assign a sample to the class whose representation is the nearest to it. 
	\citet{prototypical} propose to learn a prototype for every class and classify a sample by finding the nearest prototype. 
	Matching network \citep{matching}, on the other hand, first calculates the similarities between the test sample and each sample in the support set and then assigns the test sample to the most similar class. 
	%Normalized matching network we developed is actually a normalized version of vanilla matching network.
	Few-shot classification methods can directly calculate an emission score for each token. 
	However, to our best knowledge, there is no existing work that transfers the label dependencies for the few-shot sequence labeling task.
	
%	\vspace*{-1mm}
	\section{Conclusion}\label{sec:con}
	In this paper, 
	we propose a CRF-based framework for few-shot linguistic sequence labeling,  
	that integrates the prior knowledge of token similarities and label dependencies. 
	% Our framework is able to conduct sequence labeling with only a few labeled samples. 
	Within the framework, we propose the dependency transfer mechanism, 
	which transfers the prior knowledge of the label dependencies across domains with different label set.
	And we introduce the pair-wise embedding to few-shot sequence labeling for better token representation and more effective similarities modeling. 
	Experiments on slot tagging and named entity recognition tasks validate that both the dependency transfer mechanism and pair-wise embedding improve the labeling accuracy. 
	Analysis results prove the superiority of dependency transfer over transition rules.

	\bibliography{Atma_fewshot_arxiv}
	\bibliographystyle{my_bst}

\end{document}